\documentclass[fleqn,10pt]{wlscirep}
\usepackage[utf8]{inputenc}
\usepackage[T1]{fontenc}
\usepackage{subcaption}
\usepackage{cleveref}
\usepackage{bm}
\usepackage{tabularx}
\usepackage{multirow}
\usepackage{graphicx}
\usepackage{subcaption}
\usepackage{acro}
\usepackage{soul}
\usepackage{enumitem}

\def\QN{\text{Q-Net}}
\def\hemo{\text{hemochromatosis}}

\def\mRes{\text{ResNet}}
\def\mSWA{\text{ResNetSWA}}

\def\ilm{\text{Image-Level\ Model}}
\def\slm{\text{Scan-Level\ Model}}
\def\res{\text{ResNet18}}
\def\x{\bm{x}}
\def\X{\bm{X}}
\def\h{\bm{h}}
\def\b{\bm{b}}
\def\c{\bm{c}}
\def\f{\bm{f}}
\def\i{\bm{i}}
\def\o{\bm{o}}

\acsetup{first-style=short}
\DeclareAcronym{qsm}    {short= QSM,     long= Quantitative Susceptibility Mapping}
\DeclareAcronym{fc}     {short= FC,      long= Fully Connected}
\DeclareAcronym{bilstm} {short= Bi-LSTM, long= Bidirectional Long Short Term Memory}
\DeclareAcronym{dnn}    {short= DNN,     long= Deep Neural Network}
\DeclareAcronym{ml}     {short= ML,      long= Machine Learning}
\DeclareAcronym{mr}     {short= MR,      long= Magnetic Resonance}
\DeclareAcronym{mri}    {short= MR,      long= Magnetic Resonance Imaging}
\DeclareAcronym{swa}    {short= SWA,     long= Stochastic Weight Averaging}
\DeclareAcronym{gap}    {short= GAP,     long= Global Average Pooling}
\DeclareAcronym{cnn}    {short= CNN,     long= Convolutional Neural Network}
\DeclareAcronym{bn}     {short= BN,      long= Batch Normalization}
\DeclareAcronym{relu}   {short= ReLU,    long= Rectified Linear Unit}
\DeclareAcronym{resblk} {short= ResBlock,long= residual convolution blocks}

\title{$\QN$:  A Quantitative Susceptibility Mapping-based Deep Neural Network for Differential Diagnosis of Brain Iron Deposition in Hemochromatosis}

\author[1]{Soheil Zabihi}
\author[2]{Elahe Rahimian}
\author[3]{Soumya Sharma}
\author[4]{Sean K. Sethi}
\author[4]{Sara Gharabaghi}
\author[1,5]{Amir Asif}
\author[4,6]{E. Mark Haacke}
\author[3]{Mandar S. Jog}
\author[1,2]{Arash Mohammadi}

\affil[1]{Electrical and Computer Engineering, Concordia University, Montreal, QC, Canada}
\affil[2]{Concordia Institute for Information Systems Engineering, Montreal, QC, Canada}
\affil[3]{London Movement Disorders Centre, London Health Sciences Centre, Western University, London, ON, Canada}
\affil[4]{SpinTech MRI, Bingham Farms, MI, United States}
\affil[5]{Electrical Engineering and Computer Science, York University, Toronto, ON, Canada}
\affil[6]{Department of Radiology, Wayne State University, Detroit, MI, United States}

\begin{abstract}
Brain iron deposition, in particular deep gray matter nuclei, increases with advancing age. Hereditary Hemochromatosis (HH) is the most common inherited disorder of systemic iron excess in Europeans and recent studies claimed high brain iron accumulation in patient with Hemochromatosis. In this study, we focus on Artificial Intelligence (AI)-based differential diagnosis of brain iron deposition in HH via Quantitative Susceptibility Mapping (QSM), which is an established Magnetic Resonance Imaging (MRI) technique to study the distribution of iron in the brain. Our main objective is investigating potentials of AI-driven frameworks to accurately and efficiently differentiate individuals with Hemochromatosis from those of the healthy control group. More specifically, we developed the Q-Net framework, which is a data-driven model that processes information on iron deposition in the brain obtained from multi-echo gradient echo imaging data and anatomical information on T1-Weighted images of the brain. We illustrate that the Q-Net framework can assist in differentiating between someone with HH and Healthy control (HC) of the same age, something that is not possible by just visualizing images. The study is performed based on a unique dataset that was collected from $52$ subjects with HH and $47$ HC. The Q-Net provides a differential diagnosis accuracy of $83.16$\% and $80.37$\% in the scan-level and image-level classification, respectively.
\end{abstract}

\begin{document}
\flushbottom
\maketitle
\thispagestyle{empty}

\section*{Introduction}\label{sec:intro}
\begin{figure}[t!]
\centering
\begin{tabular}{ccc}
\subcaptionbox{\ac{qsm}\label{1}}{\scalebox{1}[-1]{\includegraphics[width = 2in]{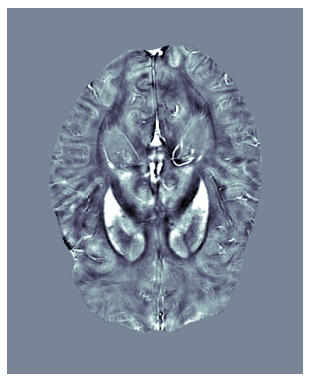}}} &
\subcaptionbox{R2*\label{2}}{\scalebox{1}[-1]{\includegraphics[width = 2in]{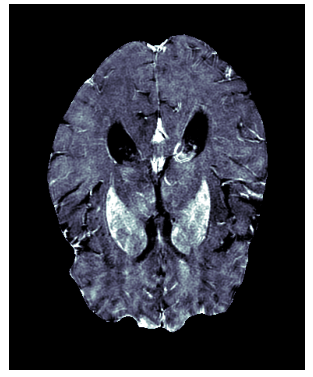}}} &
\subcaptionbox{T1W\label{3}}{\scalebox{1}[-1]{\includegraphics[width = 2in]{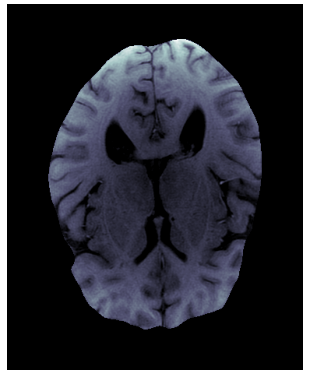}}} \\
\subcaptionbox{\ac{qsm}\label{4}}{\scalebox{1}[-1]{\includegraphics[width = 2in]{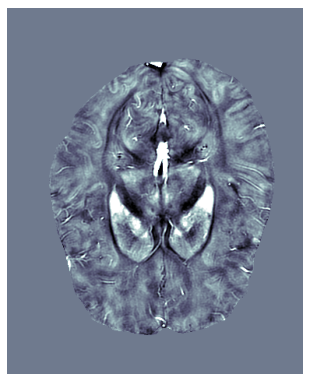}}} &
\subcaptionbox{R2*\label{5}}{\scalebox{1}[-1]{\includegraphics[width = 2in]{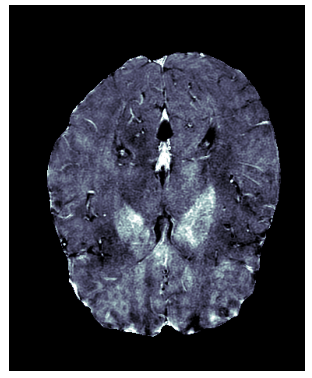}}} &
\subcaptionbox{T1W\label{6}}{\scalebox{1}[-1]{
\includegraphics[width = 2in]{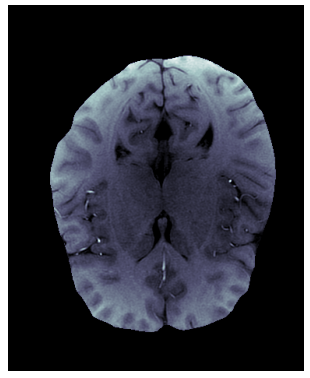}}} \\
\end{tabular}
\caption{\small Comparison of healthy control and patient with Hereditary Hemochromatosis scans. The first row shows three scans (\ac{qsm}, R2*, and T1W, respectively from left to right) associated with individuals diagnosed with Hemochromatosis. The second row shows similar scans but from a healthy individual. It can be seen that the two sets are fairly similar making it hard for the naked eye to distinguish between a healthy and unhealthy individual. Despite such similarities, the proposed approach using \ac{ml}, in particular, \ac{dnn}s perform reasonably well in such a difficult and challenging differential analysis task.
}
\label{fig:smaples}
\end{figure}

Hereditary Hemochromatosis (HH) is an inherited disorder characterized by excessive iron absorption leading to organ iron deposition~\cite{hemo2004, hemo2000}. Recent evidence suggests that brain iron deposition occurs in HH, in particular in the deep gray matter (DGM) nuclei. In this study, we propose that an Artificial Intelligence (AI)-based model, which processes information on iron deposition in the brain obtained from multi-echo gradient echo imaging data and anatomical information on T1-Weighted images of the brain, can accurately and efficiently differentiate individuals with HH from those of the Healthy Control (HC) group. In this context, we focus on Quantitative Susceptibility Mapping (QSM)~\cite{Haacke2015, Gharabaghi2020}, which is an established Magnetic Resonance Imaging (MRI) technique to study the distribution of iron in the brain. QSM is obtained from phase information acquired using gradient echo imaging sequences representing the variations of the magnetic field, which occur as a consequence of an object's magnetization within an external magnetic field~\cite{Haacke2015}. 

In this work, we investigate brain iron deposition in Hemochromatosis. The hypothesis is that the amount of iron deposition in the brain of individuals diagnosed with Hemochromatosis is different than the HCs. For example, Fig.~\ref{fig:smaples} shows three images, i.e., \ac{qsm}, R2*, and T1-Weighted (T1W),  associated with an HC and a patient with HH.  From the comparison of HC and HH's scans in Fig.~\ref{fig:smaples}, it can be seen that these two sets of images are fairly similar, which makes it difficult to distinguish them visually. Despite such similarities, we hypothesize that an advanced AI-based model can perform such differential analysis based on the distribution of iron deposition in the brain obtained from QSM and R2* sequences rather than mean values of susceptibility or R2*. The main objective of this work was to investigate and illustrate the potential of using Machine Learning (ML), in particular, Deep Neural Network (DNN) architectures, to perform this difficult and challenging classification task. More precisely, we aim to use the potential power of DNNs for end-to-end extraction of discriminative features to differentiate between HCs from HHs based on QSM, R2*, and T1W images. To the best of our knowledge, this is the first study that develops a DNN-based model to differentiate patients with HH from HCs based on QSM and R2* sequences. 

\noindent
In summary, the paper makes the following key contributions:
\begin{itemize}[noitemsep]
\item Assessing the potential of AI-based models for differentiating HCs from patients with HH through investigating iron deposition in the brain.
\item Development of a novel data-driven architecture, the $\QN$, is developed and trained over a unique dataset of QSM, R2*, and T1W images. This dataset is collected from $99$ individuals, i.e., $47$  HCs and $52$ HHs, for the purpose of studying iron deposition in the brain. 
\item The processing pipeline of the $\QN$ is an image embedding network and a sequential modeling architecture, which together form a classification paradigm for differential diagnosis. This architecture trains on its own to examine the images for extraction of expressive features, i.e., without relying on manually segmented Region-of-Interests (ROIs). More specifically, the training procedure of the $\QN$ consists of the following two stages: (i) Training an image embedding network, which is constructed based on CNNs for the purpose of dimensionality reduction of the MR images through extracting differentiating features, and; (ii) Training a Recurrent Neural Network (RNN) to accurately distinguish and classify MR images by modeling the sequential nature of the MR data.
\end{itemize}
%

\section*{Methods}\label{sec:met}

\subsection*{Dataset}\label{sec:data}
\begin{figure}
    \centering
    \scalebox{1.0}{ 
    \begin{tabular}{cc} 
        \subcaptionbox{Age-frequency plot
            \label{2}}{\includegraphics[width = 4.5in]{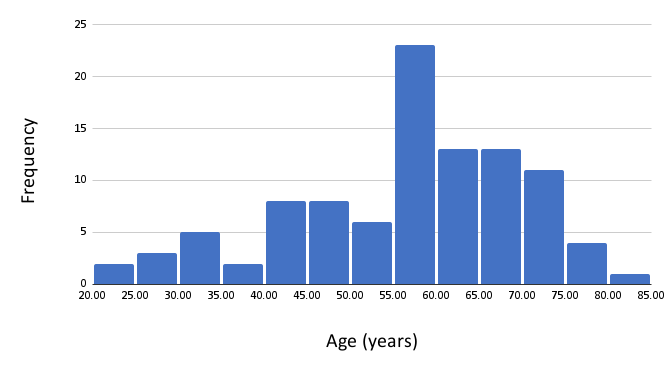}} &
        \subcaptionbox{Gender pie chart
            \label{2}}{\includegraphics[width = 2.25in]{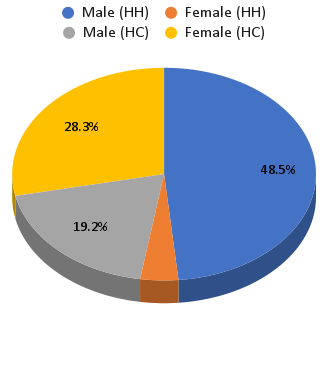}} \\
    \end{tabular}
}
\caption{\small The dataset was collected from $67$ males and $32$ females aged $20$ to $85$ years. (a) Age-frequency plot in the dataset. (b) Gender distribution in the $47$ HC and $52$ HH populations.}
    \label{fig:data_plots}
\end{figure}

The utilized dataset was collected from a total of $52$ subjects with HH and $47$ HCs from a single tertiary referral Centre between February 2019 to February 2020. The study was approved by the local Ethics committee of Health Sciences Research Ethics Board, Western University, London, Ontario (protocol number 111467). All participants gave written informed consent. Subjects with HH without liver disease were recruited from the outpatient Gastroenterology Clinic at the Movement Disorders Centre. HCs were age and sex-matched to subjects with Hereditary Hemochromatosis and were recruited by means of flyers. The age distribution of the individuals who participated in the data collection is shown in Fig.~\ref{fig:data_plots}(a). The dataset was collected from $67$ males and $32$ females aged $20$ to $85$ years. The gender distribution of the dataset for the HC and HH populations is illustrated in Fig.~\ref{fig:data_plots}(b).

A multiple echo, multiple flip angle set of gradient echo imaging data were collected. Using the STrategically Acquired Gradient Echo (STAGE) data, it was possible to create QSM, R2*, T1, and spin density maps~\cite{Haacke2015, Chen2018, Gharabaghi2020}. More precisely, STAGE uses T1W  (high FA) and proton density (low FA) data to provide multiple contrasts and quantitative maps. QSM and R2* are well-established and sensitive methods for quantifying iron in the brain, and therefore we used both of them as inputs for the model. In addition, we used T1W images as a source of information about the anatomical structure of the brain.
All subjects were imaged using a $3$T Voyager with a $32$-channel head coil (Discovery MR$750$; GE Medical Systems, Milwaukee, Wisconsin). Images were collected with the following parameters using a spoiled multi-echo Gradient Recalled Echo sequence: $TE_{1}=5 ms$, $TE_{6}=30 ms$, $\Delta TE=5 ms$, $TR = 36 ms$, $FOV = 220 mm \times 220 mm$, matrix size $366 \times 366$, and slice thickness is $2 mm$. Images were interpolated to a display resolution $0.43 mm \times 0.43 mm$.
R2* maps were generated using a previously-established method~\cite{Chen2018}. While QSM data were reconstructed for each echo individually using an in-house algorithm (SMART v2·0, MRI Institute for Biomedical Research, Bingham Farms, MI, United States) with the following steps:
(i) The Brain Extraction Tool (BET)~\cite{Smith2002} was used to isolate the brain tissue (threshold = $0.2$, erode = $4$ and island = $2000$) using the $2^{nd}$ echo where the signal intensity is highest.
(ii) The $3$D Phase Unwrapping Algorithm (3DSRNCP)~\cite{Rahman2007} was applied to unwrap the original phase data.
(iii) The Sophisticated Harmonic Artifact Reduction (SHARP)~\cite{Schweser2011} was utilized to remove unwanted background fields (threshold = $0.05$ and deconvolution kernel size = $6$).
(iv) A truncated k-space division (TKD) based inverse filtering technique (threshold = $0.1$) with an iterative approach (iteration threshold = $0.1$ and number of iterations = $4$) was used to reconstruct the susceptibility map~\cite{Tang2012}.
(v) Finally, the result susceptibility map was constructed from the QSMs from $TE_2 - TE_6$ using a method which averages each $TE$ based on its SNR~\cite{Gharabaghi2020}.

\subsection*{Data Preparation}\label{sec:dataPrep}

\vspace{.025in}
\noindent
\textbf{\textit{$\QN$ Inputs}}: Currently in the medical imaging domain, \ac{qsm} and R2* are the state-of-the-art methods to assess the regional iron concentrations. Both techniques reveal similar correlations with iron and are effective modalities for studying iron deposition. Consequently, for the design of the proposed $\QN$ framework, we use these techniques to retrieve information about regional iron concentrations along with T1W images to account for information associated with the anatomical structure of the brain. More specifically, \ac{qsm}, R2*, and T1W images are concatenated in a channel-wise fashion (resulting in a 3D input similar to RGB images) to be fed to the proposed $\QN$  architecture. It is worth noting that, in this study, from the entire scans of each individual, we consider slices that include the basal ganglia region. In other words, based on the experts' input, slices that include the basal ganglia region were selected and provided as input to the $\QN$.

\begin{figure}[t!]
\centering
\includegraphics[width=1\linewidth]{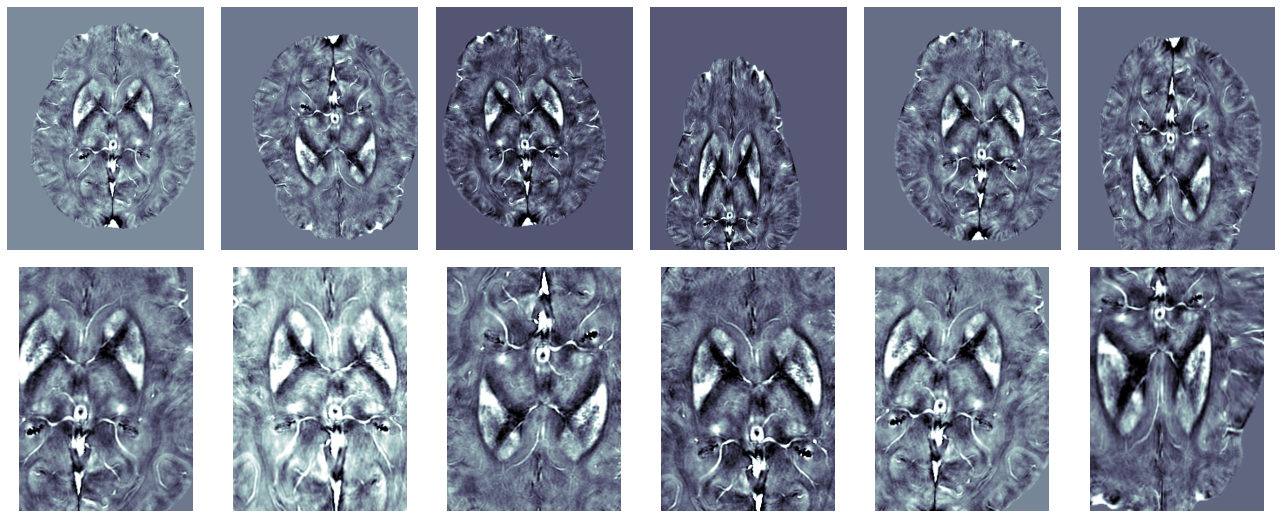}
\caption{\small Sample outputs of the data augmentation pipeline to create new images from an identical \ac{qsm} image. The first row shows augmented images of a full-size image. The second row represents the outputs of the data augmentation pipeline on a cropped region of basal ganglia. Each row shows the possible outputs of the data augmentation pipeline applied on a full-size image and cropped image. The randomness nature of the  augmentation pipeline (e.g., such as different brightness, rotation, distortion) can be observed from these images. It is worth noting that using different brightness in the augmentation pipeline (during the training phase) helps model generalizability and prevents overfitting. All the images are resulted from the same image.}
\label{fig:dataAug}
\end{figure}

\vspace{.025in}
\noindent
\textbf{\textit{Data Augmentation}}: Different data augmentation strategies are applied to the input images prior to being fed into the $\QN$ network. Generally speaking, \ac{dnn} models are data-hungry requiring a large number of input data for training purposes. Data augmentation is an effective and common strategy~\cite{Albumentations2020} to improve the performance of \ac{dnn}s in scenarios where the training set is relatively small (which is a common situation in the medical domain and this study is not an exception). There are several benefits associated with data augmentation among which is generating the extra samples in an effective and inexpensive manner.
Moreover, it helps to the generalizability of the models by avoiding overfitting. More specifically, data augmentation helps the Q-Net learn the main concepts of the underlying problem, instead of memorizing the entire dataset, i.e., overfitting, which leads to a less efficient model. Our augmentation pipeline for training consists of the following methods: Histogram Stretching ($p = 1$), Horizontal Flip ($p = 0.5$), Vertical Flip ($p = 0.5$), Random Brightness and Contrast ($p = 0.7$), Random Gamma ($p = 0.3$), Grid Distortion ($p = 0.25$), Shifting-Scaling-Rotating ($p = 0.5$), and Cropping ($p = 1$). The numbers in parenthesis represent the probability of applying the corresponding augmentation method. Fig.~\ref{fig:dataAug} illustrates sample outputs resulting from the data augmentation pipeline. During the test phase, we only use Histogram Stretching and Cropping, both with the probability of $1$.

\subsection*{The $\QN$ Framework}\label{sec:model}
\begin{figure}[t!]
\centering
\includegraphics[width=1\linewidth]{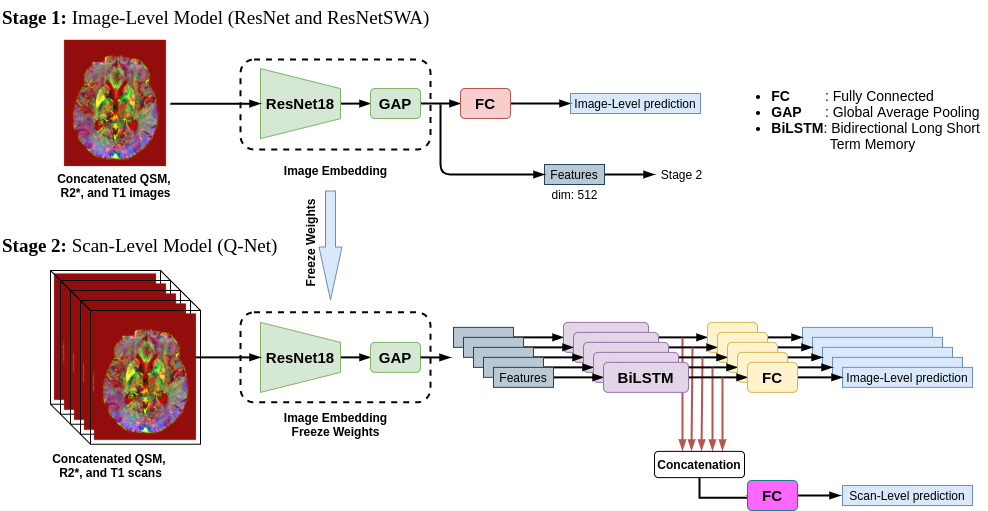}
\caption{\small The architecture of the proposed {\QN} model. The main purpose of training  $\ilm$ (Stage 1) is to build an image embedding that can be used as a feature extractor for the $\slm$ (Stage 2). The weights of the image embedding are frozen during the training phase of Stage 2. The inputs of the models are channel-wise concatenated \ac{qsm}, R2*, and T1W images. In Stage 1, input images are considered completely independent, while, in Stage 2, the entire scan associated with an individual is used via a \ac{bilstm} layer followed by two \ac{fc} layers to perform image-level and scan-level predictions.}
\label{fig:Arc}
\vspace{-.1in}
\end{figure}

\subsubsection*{Internal Architecture of the $\QN$}
Fig.~\ref{fig:Arc} shows the architecture of the proposed DNN-based $\QN$ model, which consists of two stages, i.e., Image-Level Modeling (referred to as Stage 1) and \slm ing (referred to as Stage 2). In the Image-Level Model, {\res}~\cite{resnet2016} is utilized as the backbone, which is followed by a \acl{fc} (\ac{fc}) layer. {\res} is a Convolutional Neural Network (\ac{cnn}) model composed of $18$ layers. More specifically, the utilized {\res} architecture in this study starts with a ($7\times 7$) convolution layer followed by a Batch Normalization~\cite{bn2015} (\ac{bn}) layer, Rectified Linear Unit~\cite{relu2010} (\ac{relu}) activation function, and max-pooling layer. Then, it consists of $8$ residual convolution blocks (\ac{resblk}s) with the same structure. Each \ac{resblk} contains two ($3\times 3$) convolution layers, each followed by a \ac{bn} layer, while only the first convolution has a \ac{relu} activation function. The input of each \ac{resblk} is added to its output, this is why it is referred to as a Residual block. It is worth noting that we have down-sampling before the residual connection at $3$rd, $5$th, and $7$th \ac{resblk}, which are implemented by a ($1\times 1$) convolution layer with the stride of size $2$ followed by a \ac{bn} layer. Down-sampling layers double the feature maps' dimension ($d$) while reducing its special dimensions, i.e., the width ($w$) and height ($h$), by a  factor of $2$. More specifically, the dimension of feature maps of the first convolution layer is $64$, which increases to $128$, $256$, and $512$ after each down-sampling. At the end, the final layers of the $\res$ architecture are consist of a \acl{gap}~\cite{gap2013} (\ac{gap}) layer followed by a \ac{fc} layer. Although the \ac{gap} layer, similar to the max-pooling layer, is used to reduce the spatial dimensions of the feature maps, it performs an extreme type of dimensionality reduction, where feature maps' tensor with dimension $h\times w\times d$ are reduced in size to a lower-dimensional space of $1\times 1\times d$. In other words, \ac{gap} layer reduces each $h\times w$ feature map to a single number by simply taking the average of all $h \times w$ values of that feature map.

The main purpose of the first stage is to build an image-level embedding network. To achieve this objective, the utilized {\res} architecture is initialized with ImageNet~\cite{imagenet2009} weights, and the whole network was then trained based on the in-house dataset. As stated previously, \ac{qsm}, R2*, and T1W, were concatenated in a channel-wise fashion and fed to Stage 1. The Stand-alone Image-Level Model cannot completely capture the underlying characteristics of input sequences as images from the same scan are considered completely independent from one another. Capitalizing on the fact that all images of the same scan/patient are potentially correlated, Stage 2, i.e., {\slm}, was introduced to use and model inherit correlations among images of a patient. The {\slm} utilized the trained embedding network of Stage 1 as an image feature extractor to feed a one-layer Bidirectional Long-Short Term Memory~\cite{bilstm1997} (\ac{bilstm}) network. The latter was utilized to combine the extracted features of the entire images of a patient. More specifically, we used the extracted features from each image after the \ac{gap} layer in the \ilm's architecture. Then, these features were used to train the \ac{bilstm} layer. Intuitively speaking, the \ac{bilstm} module integrates extracted features of all images of a patient.

Given $\X=[\x^1, \x^2, \dots, \x^t, \dots, \x^{T}]$ as an input sequence with length of $T$, where $\x^t = [x^t_1, x^t_2, \dots, x^t_d]$ is a $d$-dimensional feature vector. The feature vector is extracted from input MRI sequence at position $t$ representing the image index in the entire scan of an individual. The internal structure of an Long-Short Term Memory~\cite{lstm1997} (LSTM) cell is represented as follows
\begin{eqnarray}
\text{Input Gate}:  \i^t &=& \sigma (W_i \x^t + U_i \h^{t-1} + \b_i), \label{eq:1}\\
\text{Output Gate}:  \o^t &=& \sigma (W_o \x^t + U_o \h^{t-1} + \b_o), \label{eq:2}\\
\text{Forget Gate}:  \f^t &=& \sigma (W_f \x^t + U_f \h^{t-1} + \b_f), \label{eq:3}\\
\text{Cell State}:  \c^t &=& \f_t \circ \c^{t-1} + \i^t \circ \tanh(W_c \x^t + U_c \h^{t-1} + \b_c), \label{eq:4}\\
\text{Cell Output}:  \h^t &=& \o^t \circ \tanh(\c^t), \label{eq:5}
\end{eqnarray}
where term $W_*$ and $U_*$ represent weight matrices, and $\b_*$ represents the bias vectors. Furthermore, in Eqs.~\eqref{eq:1}-\eqref{eq:5}, $\h^{t}$ and $\c^{t}$ represent the hidden-state (also called cell-output) and cell-state of the LSTM, respectively. More specifically, the information flow of the LSTM's internal cell structure is controlled by a gating logic consisting of three gates: (i) \textit{Input Gate}, which determines what information based on the input $\x^t$ and the previous hidden state $\h^{t-1}$ will be forwarded to the memory cell; (ii) \textit{Output Gate}, which controls the information that will be passed to the next hidden state $\h^{t}$, and; (iii) the \textit{Forget Gate} to determine the required information from the prior cell-state $\c^{t-1}$. Terms $\sigma(\cdot)$ and $\tanh(\cdot)$ denote the sigmoid and tangent hyperbolic activation functions, respectively; and operator ``$\circ$'' denotes element-wise multiplication of two vectors. However, the LSTM cell can only capture previous context. To overcome this limitation, in \ac{bilstm} architecture, $2$ LSTMs are used in parallel to process the input in forward and backward directions. In other words, the input sequence is fed in normal time order for one network, and in the reverse time order for the other. Processing input in opposite directions increases the amount of available information for the model. In the proposed $\QN$ model, the combined features, resulted from the \ac{bilstm} module, were utilized by two independent \ac{fc} layers to perform scan-level and image-level predictions.

\subsubsection*{Training Procedure}
The training procedure of the $\QN$ consists of the following two steps: (i) Training the Image-Level Model (Stage 1) to be used as an ``\textit{Image embedding}'', and; (ii) Training the Scan-level Model (Stage 2) to perform scan-level and image-level predictions. In this context, the training step of Stage 1 is performed to generate an ``\textit{Image embedding}'', which can be seen as a function that takes a 3D input (channel-wise concatenated \ac{qsm}, R2*, and T1W images) and returns a numeric vector as its representation. In other words, the image embedding translates/converts high-dimensional input images into numeric vectors in a lower-dimensional space. The embedding model, therefore, allows effective implementation of \ac{ml} models on our large dimensional dataset by converting them into low-dimensional vectors. During the training phase of Stage 2 (scan-level modeling), image embedding from Stage 1 is used as a feature extractor to predict both image and scan-level labels.  Image embedding is constructed by removing the \ac{fc} layer from the architecture of Stage 1. It is worth noting that the embedding's weights are not updated during the training phase of Stage 2.

For training the architecture of both stages, cross-entropy loss is used. We used $10$-fold cross-validation for model training and evaluation. More specifically, we split the dataset into $10$ folds, where each fold consists of $10$ individuals. By considering each fold as a test set, we train the model on the remaining $9$ folds. Therefore, $10$-fold cross-validation allows validation of the $\QN$ on the entire dataset.

\vspace{.025in}
\noindent
\textit{Learn an Image Embedding}: For training an image embedding, we have to train an image-level classifier, i.e., identifying a given slice as belonging to a healthy or unhealthy category. For this purpose, we used {\res}~\cite{resnet2016} architecture as the backbone of the model, which is  initialized with ImageNet~\cite{imagenet2009} pre-trained weights. More specifically, we use the Transfer Learning (TL) technique to update the parameters of the model while the weights of the ResNet18 backbone are not frozen during the first training stage. Using TL has a considerable impact on performance improvement as the model leverages the capacity of DNN algorithms to learn discriminant features from large datasets. Moreover, during the training phase of the Image-Level Model, Stochastic Weight Averaging~\cite{swa2018} (\ac{swa}) technique is applied to further improve the results. More specifically, \ac{swa} uses a modified learning rate schedule to compute an equal average of the weights traversed by Stochastic Gradient Decent (SGD). Utilization of the \ac{swa} technique can potentially improve the overall achievable performance compared to conventional training mechanisms~\cite{swa2018}. For the first stage, the duration of training is $100$ epochs with a batch size of $64$, where Adam optimizer~\cite{adam2014} is used for model optimization. In first stage of the training, the learning rate starts from $3e-4$ and declines to $1e-4$ with cosine annealing~\cite{sgdr2017}.

\vspace{.025in}
\noindent
\textit{Sequential Modeling}: To account for potential correlations between consecutive \ac{qsm}, R2*, and T1W images in an individual's MRI series, Recurrent Neural Network (RNN) architectures  (a class of \ac{dnn}s for sequential modeling) is utilized within the $\QN$ framework. However, due to the curse of dimensionality, direct use of images for training RNNs is computationally expensive, and it needs a large amount of data. To address these problems, we used our pre-trained image embedding network resulted from Image-Level Model. As shown in Fig.~\ref{fig:Arc}, for the embedding network, we use the Image-Level Model and remove its final \ac{fc} layers. More specifically, we used the extracted features from each image after the \ac{gap} layer in Image-Level Model's architecture. Then, the information of these extracted features, i.e., embedded images, are combined using \ac{bilstm} layer. Therefore, with these considerations, the Scan-Level Model has the capacity of investigating the entire images of an individual's scan, while considering their correlation leading to more accurate results.

In the context of sequential modeling, we used the following two strategies: (i) Image-Level Classification, and; (ii) Scan-Level Classification. In the first strategy, we classified each slice but unlike the training strategy used in Stage 1 (Image-Level Model), the information of slices in each scan are integrated by a \ac{bilstm} layer, i.e., they are not considered as independent images. Then, a \ac{fc} layer (shown as yellow blocks in Fig.~\ref{fig:Arc}) is leveraged to classify each slice. In the second strategy, i.e., the Scan-Level Classifier, the output features of the \ac{bilstm} layer are concatenated and a \ac{fc} layer is utilized to merge the information of all slices to predict the scan-level labels. The duration of training is $50$ epochs with a batch size of $64$, and the learning rate is fixed to $3e-4$. It should be noted that the weights of the embedding network are frozen during this training stage. When a network is frozen, it means that its weights cannot be modified further.

\section*{Results}

\subsection*{Performance Evaluation based on Input Configuration}
\begin{figure}[t!]
    \centering
    \scalebox{1.0}{
    \begin{tabular}{c}
        \includegraphics[width = 3.4in]{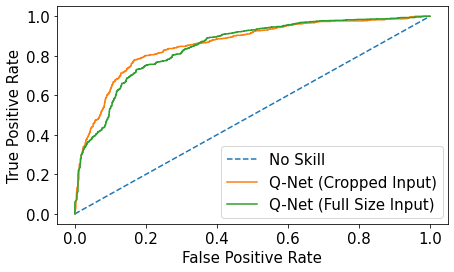}
    \end{tabular}
    }
\caption{ \small ROC curve comparison for full-size image and cropped-image inputs for the image-level classification. The points along the diagonal-dashed line indicate a random (No Skill)  classifier, which is plotted as a baseline.
The AUCs of the {\QN} model for full-size and cropped input images are $0.851$ and $0.866$, respectively.
}\label{fig:ROC2}
\end{figure}
The following two scenarios are considered to evaluate the effects of different inputs on the performance of the proposed $\QN$: (i) The full-size image is utilized as the input of the model, and; (ii) Each image is cropped to fit the basal ganglia region, which is then used as the input to the model. In both scenarios, \ac{qsm}, R2*, and T1W images are concatenated in a channel-wise fashion resulting in 3D inputs. In both scenarios, four different metrics, i.e., Accuracy, Sensitivity, Specificity, and F1-score are utilized for performance evaluations. Results are shown in Table~\ref{table1}. The results reported in Table~\ref{table1} are based on $10$-fold cross-validation.

As shown in Table~\ref{table1}, we have nearly equivalent accuracy and F1-score in all models for full-size image and cropped image inputs. Therefore, it could be said that we have nearly equal precision and recall, which is a piece of evidence that all models have the same performance in both classes, i.e., HC and HH prediction. In other words, models are not biased to predict a specific class more often than another. As shown in Table~\ref{table1}, by feeding cropped images to the models, we have an improvement in all criterion except Specificity. In other words, models are more capable to identify healthy individuals when full-size images are used as the input, which is also depicted in Fig.~\ref{fig:ROC2}, where {\QN} with full-size image inputs occasionally score better in a specific region than {\QN} with cropped image inputs. However, in practice, the {\QN} with cropped image inputs have higher AUC ($0.866$) than {\QN} with full-size image inputs ($0.851$), which means it performs better in general.

\subsection*{Performance Evaluation from ``Modeling'' Perspective}
\begin{table*}[t!]
\caption{\small Results of the Image-Level classifiers for full-size and cropped image inputs. $10$-fold cross-validation is utilized for evaluating models on the entire dataset. The reported values are shown in MEAN $\pm$ STD format.}
    \label{table1}
    \centering
    \scalebox{0.883}{  
    \begin{tabular}{ |c|cccc|cccc|}
        \hline
        Input &
        \multicolumn{4}{|c|}{\textbf{Full Size Image}}&
        \multicolumn{4}{|c|}{\textbf{Cropped Image}}
        \\ \hline
        Metric &
        Accuracy & Sensitivity & Specificity & F1-score &
        Accuracy & Sensitivity & Specificity & F1-score
        \\ \hline
        \mRes &
        $68.19 \pm 0.24$ & $67.46 \pm 1.24$ & $69.00 \pm 1.20$ & $68.21 \pm 0.26$ &
        $68.46 \pm 0.26$ & $69.62 \pm 1.28$ & $67.16 \pm 1.39$ & $68.47 \pm 0.29$
        \\ 
        \mSWA &
        $70.36 \pm 0.20$ & $70.44 \pm 0.39$ & $70.26 \pm 1.20$ & $70.38 \pm 0.21$ &
        $71.17 \pm 0.13$ & $77.55 \pm 0.12$ & $64.06 \pm 0.68$ & $71.01 \pm 0.14$
        \\ 
        \QN &
        $75.47 \pm 0.80$ & $70.36 \pm 1.77$ & $80.91 \pm 1.54$ & $75.43 \pm 0.79$ &
        $80.37 \pm 0.70$ & $82.43 \pm 1.76$ & $78.16 \pm 1.14$ & $80.35 \pm 0.70$
        \\ \hline
    \end{tabular}
    }
\end{table*}
\begin{figure}[t!]
    \centering
    \scalebox{1.0}{ 
    \begin{tabular}{cc} 
        \subcaptionbox{ROC curves for full-size input images
            \label{2}}{\includegraphics[width = 3in]{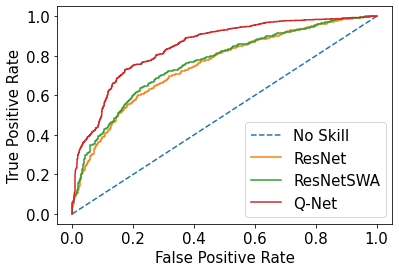}} &
        \subcaptionbox{ROC curves for cropped input images
            \label{2}}{\includegraphics[width = 3in]{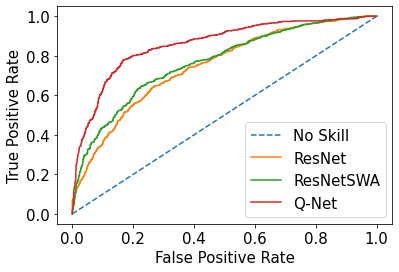}} \\
    \end{tabular}
}
\caption{\small ROC curves. The ``True Positive Rate'' is the ratio of observations that are correctly predicted from all positive observations. Likewise, the ``False Positive Rate'' is the ratio of observations that are incorrectly predicted to be positive from all negative observations. The points along the diagonal-dashed line indicate a random (No Skill)  classifier, which is plotted as a baseline. In a sense that the closer the ROC curve is to this $45$ degree diagonal line, the less accurate the model is. The AUCs of the {\mRes}, {\mSWA}, and {\QN} models are $0.747$, $0.763$, and $0.851$ in (a), and $0.749$, $0.766$, and $0.866$ in (b), respectively.}
    \label{fig:ROC1}
\end{figure}

\begin{table*}[t!]
\caption{\small DeLong test results for pairwise comparison of all ROC curves. If p-value is less than the pre-defined $5\%$ ($p < 0.05$), the conclusion is that the two compared models are different, i.e., the null hypothesis ($H_0$) will be rejected. The significance of difference is categorized in multiple level as follows:
not-significant ($ns$): $0.05 < p \leq 1$,
\quad*: $0.01 < p \leq 0.05$,
\quad**: $0.001 < p \leq 0.01$,
\quad***: $0.0001 < p \leq 0.001$,
\quad****: $p \leq 0.0001$.}
    \label{delong}
    \centering
    \scalebox{1.03}{  
    \begin{tabular}{ |ccc|cc|c|cc|c|}
        \hline
        \multicolumn{9}{|l|}{Pairwise comparison of all ROC curves:}
        \\
        &\multicolumn{5}{l}{$H_0: \textbf{AUC}_1 = \textbf{AUC}_2$}& \multicolumn{3}{l|}{\textit{$H_0$: Null Hypotheses}}
        \\
        &\multicolumn{5}{l}{$H_1: \textbf{AUC}_1 \neq \textbf{AUC}_2$}& \multicolumn{3}{l|}{\textit{$H_1$: Alternative Hypotheses}}
        \\
        \hline
        \multicolumn{1}{|c|}{\textbf{Input}}&
        \multicolumn{2}{c|}{\textbf{Paired Criterion Variables}}&
        \multicolumn{2}{c|}{\textbf{AUCs}}&
        \multicolumn{1}{c|}{\textbf{Difference}}&
        \multicolumn{2}{c|}{\textbf{Significance Test}}&
        \multicolumn{1}{c|}{\textbf{Significance}}
        \\

        \multicolumn{1}{|c|}{{}}&$\textbf{Variable}_1$ & $\textbf{Variable}_2$ &
        $\textbf{AUC}_1$ & $\textbf{AUC}_2$ &
        \textbf{Std Error} & \textbf{z-score} & \textbf{p-value} & \textbf{Level}
        \\ \hline
        \multicolumn{1}{|c|}{\textbf{Full}}&\mSWA & \mRes &
        $0.763$ & $0.747$ &
        $0.0083$ & $1.928$ & $0.0589$ & $ns$ 
        \\ 
        \multicolumn{1}{|c|}{\textbf{Size}}&\textbf{\QN} & \mRes &
        $0.851$ & $0.747$ &
        $0.0087$ & $11.954$ & $6.2e-33$ & $****$ 
        \\ 
        \multicolumn{1}{|c|}{\textbf{Image}}&\textbf{\QN} & \mSWA &
        $0.851$ & $0.763$ &
        $0.0084$ & $10.476$ & $1.1e-25$ & $****$ 
        \\ \hline

        \multicolumn{1}{|c|}{\textbf{Cropped}}&\mSWA & \mRes &
        $0.766$ & $0.749$ &
        $0.0086$ & $1.9767$ & $0.0480$ & $*$ 
        \\ 
        \multicolumn{1}{|c|}{\textbf{Image}}&\textbf{\QN} & \mRes &
        $0.866$ & $0.749$ &
        $0.0094$ & $12.447$ & $1.5e-35$ & $****$ 
        \\ 
        \multicolumn{1}{|c|}{\textbf{}}&\textbf{\QN} & \mSWA &
        $0.866$ & $0.766$ &
        $0.0089$ & $11.236$ & $2.7e-29$ & $****$ 
        \\ \hline
    \end{tabular}
    }
\end{table*}
\begin{figure}
    \centering
    \scalebox{1.0}{ 
    \begin{tabular}{cc} 
        \subcaptionbox{AUC boxplots for full-size input images
            \label{2}}{\includegraphics[width = 3in]{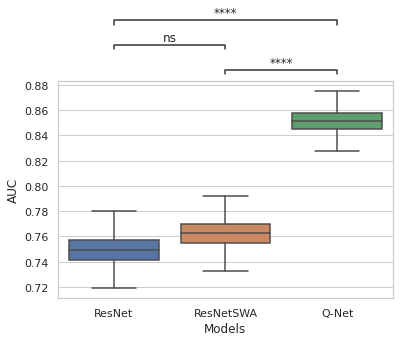}} &
        \subcaptionbox{AUC boxplots for cropped input images
            \label{2}}{\includegraphics[width = 3in]{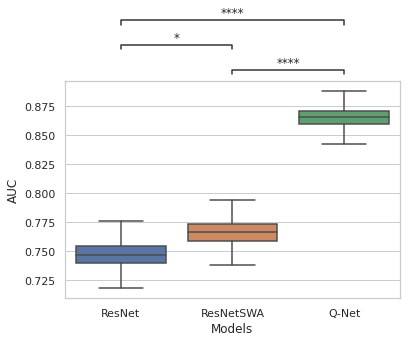}} \\
    \end{tabular}
}
\caption{\small AUC boxplots for all models, black horizontal line in IQR is the median AUC. The significance level for p-values are marked the same as Table~\ref{delong} at top of each plot as follows: not-significant ($ns$): $0.05 < p \leq 1$,
\quad*: $0.01 < p \leq 0.05$,
\quad**: $0.001 < p \leq 0.01$,
\quad***: $0.0001 < p \leq 0.001$,
\quad****: $p \leq 0.0001$.}
    \label{fig:boxplot}
\end{figure}

In addition to considering two different input configurations, the performance of the proposed $\QN$ framework is evaluated based on three different models, i.e., ``{\mRes}'', ``{\mSWA}'', and the proposed ``{\QN}''. Please note that {\mSWA} model is similar in nature to the {\mRes} with the difference that the \ac{swa} technique~\cite{swa2018} is leveraged during the training phase. As shown in Table~\ref{table1}, the performance of {\QN} is superior to that of {\mRes} and {\mSWA} models for image-level predictions. For example, {\QN} shows $7.28\%$ and $5.17\%$ accuracy improvement in comparison to {\mRes} and {\mSWA} for full-size image inputs, respectively. Moreover, these accuracy improvements exceed to $11.91\%$ and $9.2\%$ for cropped image inputs, respectively.
As shown in Table~\ref{table1}, the reported results of {\mSWA} are superior to those obtained from {\mRes}. For example, the accuracy is improved more than $2\%$ in {\mSWA} model in comparison to {\mRes} in both full-size and cropped image inputs scenarios. This illustrates that the \ac{swa} technique can lead to a much better generalization. However, the results on the $\mRes$ and $\mSWA$ are achieved on the image-level, i.e., each image is considered entirely independent from other images of the same scan, which is a problem because, in practice, specialists consider entire scan of the patient for diagnosis. The proposed $\QN$ addresses this issue by coupling a \ac{bilstm} layer with the trained $\mSWA$ as an image embedding module.

To have a better sense of the model's performance, we also provided graphical representations in Fig.~\ref{fig:ROC1} by plotting Receiver Operator Characteristic (ROC) curves, which show the diagnostic ability of binary classifiers. The ROC curve illustrates the trade-off between True Positive Rate (or Sensitivity) and False Positive Rate (or $1 -$ Specificity). As a result, a classifier that brings the ROC curve closer to the top left corner shows better performance. Therefore, as shown in the Fig.~\ref{fig:ROC1}, the {\mSWA} model shows a slight improvement compared to the {\mRes} model, while {\QN} shows a significant improvement compared to its counterparts over the two different input configurations. To summarize the performance of each classifier into a single criterion one common approach is to calculate the area under the ROC curve (abbreviated to AUC). For the full-size image inputs scenario, the AUCs of the {\mRes}, {\mSWA}, and {\QN} models are $0.747$, $0.763$, and $0.851$, respectively. For the cropped image inputs scenario, AUCs of the {\mRes}, {\mSWA}, and {\QN} models are $0.749$, $0.766$, and $0.866$, respectively. In both cases, {\QN} shows considerable improvement.


When selecting or discarding a classifier based on AUCs, it is useful to quantify the certainty of AUCs differences. The DeLong~\cite{delong} method tests whether two models have significantly different performance and accounts for the uncertainty due to the randomness of the finite training set and the evaluation on a common test set. Testing ``whether one model is significantly better than another" can be rephrased as testing the null hypothesis that two models have statistically equal AUCs against the alternative hypothesis that the statistics are unequal. Thus, the null and alternative hypotheses for the two models are  expressed as follows: $H_0: \textbf{AUC}_1 = \textbf{AUC}_2$ versus $H_1: \textbf{AUC}_1 \neq \textbf{AUC}_2$. Table~\ref{delong} shows the statistical analysis for pairwise comparison of all ROC curves by DeLong tests. If the significant threshold for p-value is less than the  $5\%$ ($p<0.05$), the conclusion is that the null hypothesis is false, i.e., the two compared models are different. In Table~\ref{delong}, the significance of the difference between two models is categorized in multiple levels as follows:
\begin{enumerate}[label=(\roman*)]
  \item $ns$ : $0.05 < p \leq 1$,
  \item *    : $0.01 < p \leq 0.05$,
  \item **   : $0.001 < p \leq 0.01$,
  \item ***  : $0.0001 < p \leq 0.001$,
  \item **** : $p \leq 0.0001$.
\end{enumerate}
where $ns$ stands for not-significance. To have a better insight into statistical test results, we also utilize the bootstrapping approach to visualize the AUCs distribution as shown in Fig.~\ref{fig:boxplot}. In this figure, the AUC distribution across all images for each model with full-size and cropped input images is shown. The boxplot for each model shows the Interquartile Range (IQR), which is based on dividing the performance of each model for all inputs into quartiles. The median AUC is shown by a horizontal line in each boxplot.

\subsection*{Scan-Level Classification}
\begin{table*}[t!]
    \caption{\small Results of the Scan-Level classifiers for full-size and cropped image inputs.}
    \label{table2}
    \centering
    \begin{tabular}{ |c|cccc|cccc|}
        \hline
        Input &
        \multicolumn{4}{|c|}{\textbf{Full Size Image}}&
        \multicolumn{4}{|c|}{\textbf{Cropped Image}}
        \\ \hline
        Metric &
        Accuracy & F1-score & Sensitivity & Specificity &
        Accuracy & F1-score & Sensitivity & Specificity
        \\ \hline
        \mRes &
        $75.79$ & $75.78$ & $77.55$ & $73.91$ &
        $76.84$ & $76.83$ & $77.55$ & $76.07$
        \\ 
        \mSWA &
        $76.89$ & $76.84$ & $80.43$ & $73.46$ &
        $81.05$ & $80.77$ & $81.83$ & $79.56$
        \\ 
        \QN &
        $80.00$ & $79.99$ & $81.63$ & $78.26$ &
        $83.16$ & $83.14$ & $85.71$ & $80.43$
        \\ \hline
    \end{tabular}
\end{table*}
\begin{figure}[t!]
    \centering
    \scalebox{1.0}{
    \begin{tabular}{c}
        \includegraphics[width = 3.4in]{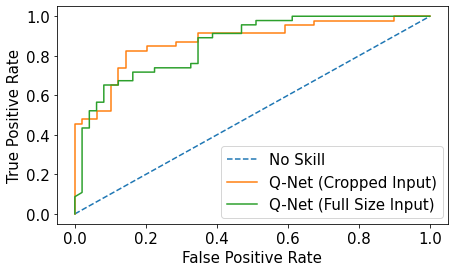}
    \end{tabular}
    }
\caption{ \small ROC curve comparison for full-size image and cropped-image inputs in scan-level classification. The points along the diagonal-dashed line indicate a random (No Skill)  classifier, which is plotted as a baseline. The AUCs of the {\QN} model for full-size and cropped input images to classify patients in two categories (HC and HH) are $0.860$ and $0.875$, respectively. }\label{fig:ROC_compare_patient}
\end{figure}

Table~\ref{table2} shows the performance of the proposed $\QN$ framework in scan-level classification based on full-size and cropped inputs configurations. As mentioned previously, {\mRes} and {\mSWA} are image-level classifiers. Therefore, in order to have scan-level predictions for these models, we used a voting mechanism. More specifically, the voting is performed as follows: Results from HC and HH predictions for all images corresponding to a patient are counted, and if the number of HC predicted images is more than HH, the patient is classified as a HC, and vice versa. On the contrary, in the {\QN}, we have a fusion module (\ac{fc} layer) that combines all extracted features resulted from the \ac{bilstm} layer to make decisions about the entire scan, as shown in Fig.~\ref{fig:Arc}. Reported results in Table~\ref{table2} show that {\QN} model with cropped image inputs performs better than other approaches. For example, accuracy of {\mRes}, {\mSWA}, and {\QN} for full-size input images are $75.79\%$, $76.89\%$, and $80\%$ respectively, while these are increased to $76.84\%$, $81.05\%$, and $\bm{83.16}\%$ for input images roughly cropped around the basal ganglia region. Moreover, in comparison to {\mRes} and {\mSWA} for full-size input images, {\QN} shows $4.21\%$ and $3.11\%$ accuracy improvements, respectively. For the cropped input images, on the other hand, improvements change to $6.23\%$ and $2.11\%$, respectively.
In comparison to {\mRes} and {\mSWA}, this improvement is achieved because of the $\QN$ model's specific characteristics, i.e., considering and integrating all images associated with a patient for making a decision. While compared to the {\QN} with full-size image inputs, the better performance of {\QN} resulted because of the extra bias in the model's inputs, i.e., cropping the input images around the ganglia region in the brain. This improvement is shown in the ROC plots illustrated in Fig.~\ref{fig:ROC_compare_patient}. The AUCs of the {\QN} model for full-size and cropped input images to classify patients in two categories (HC and HH) are $0.860$ and $0.875$, respectively.

\section*{Discussion}\label{sec:disc}
\begin{figure}[t!]
    \centering
    \begin{tabular}{cccc}
\frame{\scalebox{1}[-1]{\includegraphics[width = 1.5in]{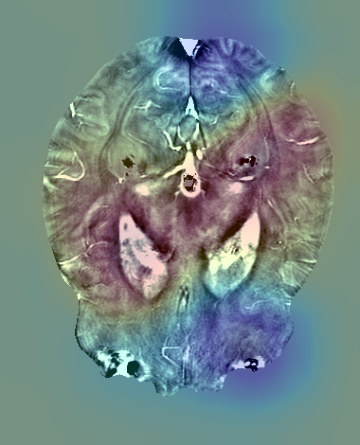}}} \!\!\!\!&\!\!\!\!
\frame{\scalebox{1}[-1]{\includegraphics[width = 1.5in]{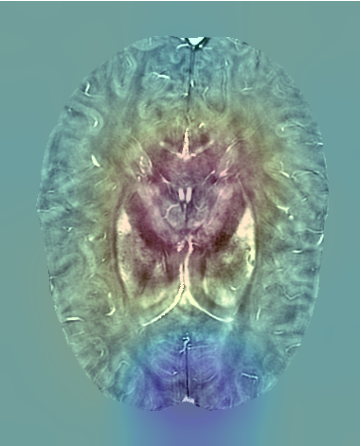}}} \!\!\!\!&\!\!\!\!
\frame{\scalebox{1}[-1]{\includegraphics[width = 1.5in]{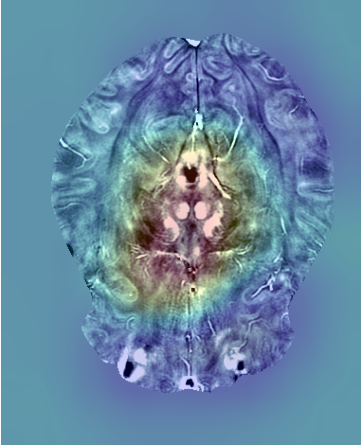}}} \!\!\!\!&\!\!\!\!
\frame{\scalebox{1}[-1]{\includegraphics[width = 1.5in]{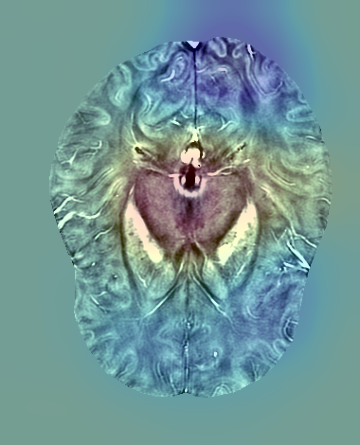}}} \\
\frame{\scalebox{1}[-1]{\includegraphics[width = 1.5in]{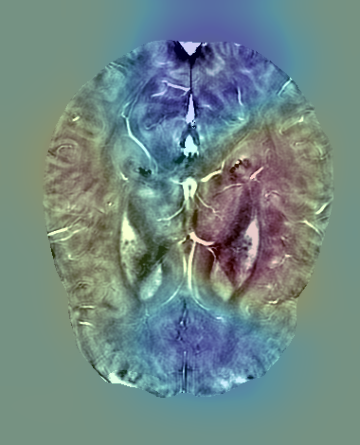}}} \!\!\!\!&\!\!\!\!
\frame{\scalebox{1}[-1]{\includegraphics[width = 1.5in]{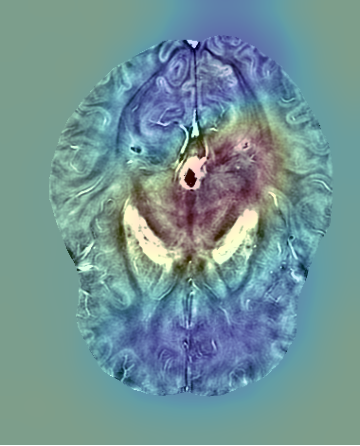}}} \!\!\!\!&\!\!\!\!
\frame{\scalebox{1}[-1]{\includegraphics[width = 1.5in]{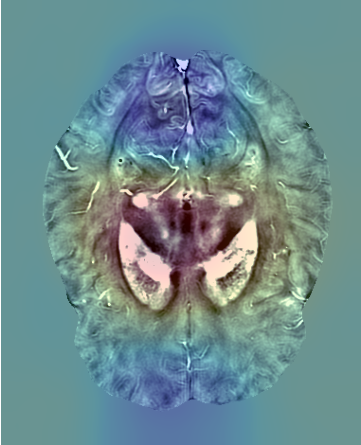}}} \!\!\!\!&\!\!\!\!
\frame{\scalebox{1}[-1]{\includegraphics[width = 1.5in]{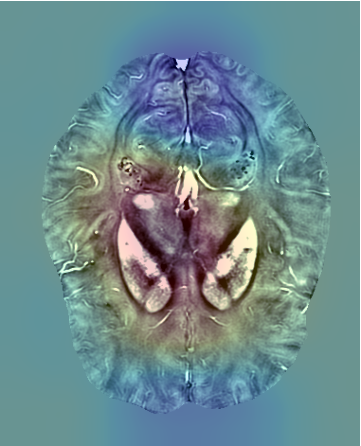}}} \\
    \end{tabular}
    \caption{Heat maps resulted from ``Class Activation Map (CAM)'' method for full-size input images. Red regions represent the part of the image that the {\QN} focuses on while trying to make a prediction.}
    \label{fig:CAM_full_size}
\end{figure}
\begin{figure}[t!]
    \centering
    \begin{tabular}{cccc}
\frame{\scalebox{1}[-1]{\includegraphics[width = 1.5in]{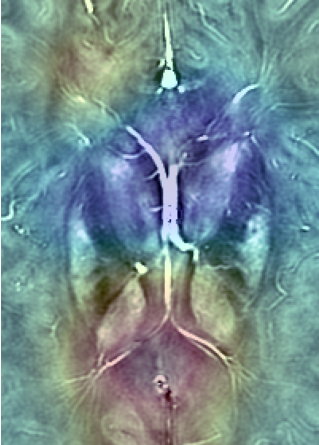}}} \!\!\!\!&\!\!\!\!
\frame{\scalebox{1}[-1]{\includegraphics[width = 1.5in]{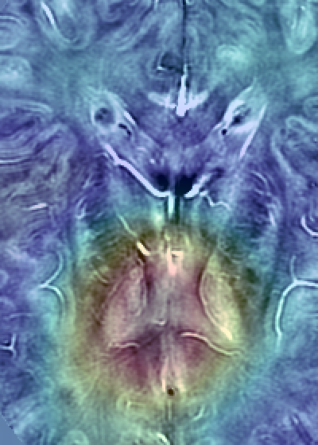}}} \!\!\!\!&\!\!\!\!
\frame{\scalebox{1}[-1]{\includegraphics[width = 1.5in]{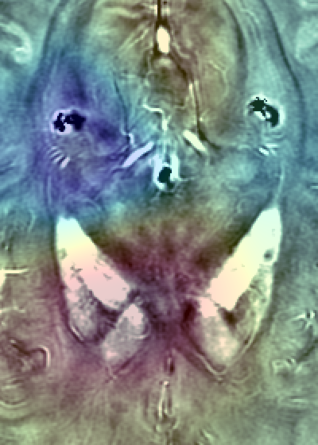}}} \!\!\!\!&\!\!\!\!
\frame{\scalebox{1}[-1]{\includegraphics[width = 1.5in]{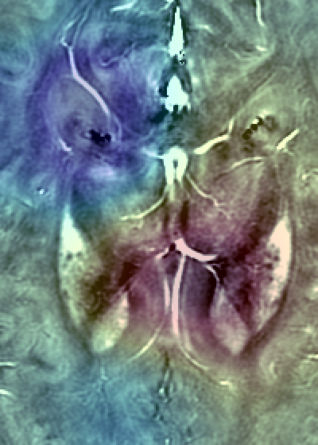}}} \\
\frame{\scalebox{1}[-1]{\includegraphics[width = 1.5in]{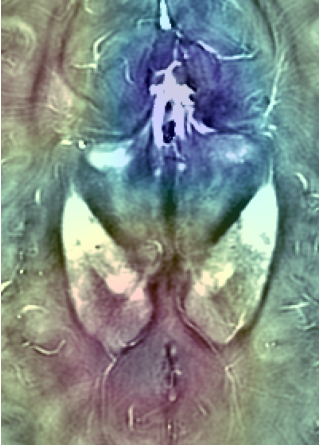}}} \!\!\!\!&\!\!\!\!
\frame{\scalebox{1}[-1]{\includegraphics[width = 1.5in]{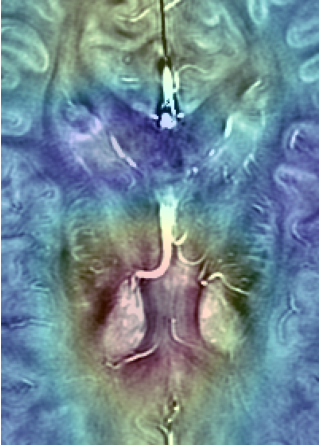}}} \!\!\!\!&\!\!\!\!
\frame{\scalebox{1}[-1]{\includegraphics[width = 1.5in]{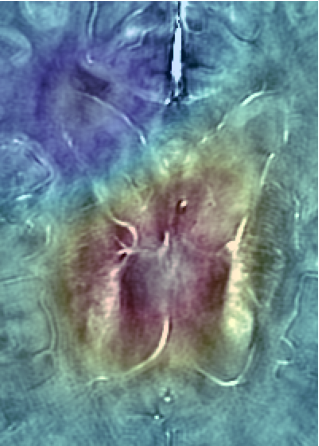}}} \!\!\!\!&\!\!\!\!
\frame{\scalebox{1}[-1]{\includegraphics[width = 1.5in]{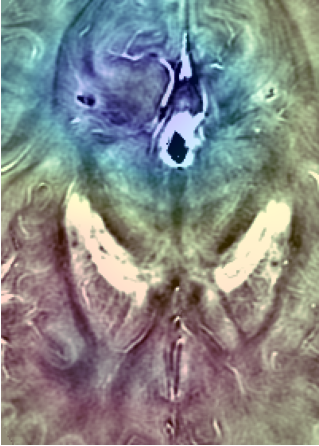}}} \\
    \end{tabular}
    \caption{Heat maps resulted from CAM method for cropped input images. Red regions represent the part of the image that the {\QN} focuses on while trying to make a prediction.}
    \label{fig:CAM_crop}
\end{figure}

Brain iron deposition increases with age, making it very difficult to differentiate between individuals with an iron overload disorder and healthy individuals by simply visualizing their MR images. Therefore, in this study, we equipped ourselves with an advanced AI model, the $\QN$, to discover the potential relationship of the {\hemo} with  changes in brain iron content. To date, there are only a few AI models developed based on QSM sequences. Here, we focus on highlighting the differences between this work and previous relevant studies. On the one hand, Lee \textit{et al.}~\cite{Lee2020} focused on classical Machine Learning (\ac{ml}) methods such as Support Vector Machine (SVM) and Logistic Regression (LR) models, to differentiate between Healthy Controls (HCs) and patients at the early stage of PD. The accuracy of their optimal SVM and LR models on a dataset collected from $52$ individuals ($24$ patients and $27$ HCs) were $79\%$ and $73\%$, respectively. One main drawback of this approach is utilizing manually segmented ROIs such as the basal ganglia, for feature extraction. This is a time-consuming approach and requires clinical expertise. Moreover, hand-crafted and pre-defined features are constructed by simply using the mean QSM values of the segmented areas in the deep gray matter. This, in turn, results in the interpretation of high dimensional QSM scans with only seven numbers, which means that important textural and special information is ignored. Bin \textit{et al.}~\cite{Haacke:QSM1} suggested a hybrid feature extraction method to differentiate individuals with PD from healthy controls (HCs). First, they manually extracted ROIs of the substantia nigra (SN) region in the brain using QSM images. Then, these ROIs were utilized to extract hand-crafted radiomics features from the SN region. Two machine methods, i.e., LR and SVM, were employed to predict PD and HC. Along with this approach, the ROIs were used to crop the QSM images to get the patches containing the SN region of the brain. These patches were then used to train a Convolutional Neural Network (CNN)-based model. Finally, they combined these two types of features, i.e., hand-crafted radiomics features and CNN features, to perform the classification task.
On the other hand, some studies focused on utilizing Deep Neural Networks (DNN) models to reconstruct QSM and improve the reconstruction process. For instance, Jo \textit{et al.}~\cite{Jo2020} proposed a DNN-based approach, i.e., modified U-Net~\cite{unet2015} structure, to improve QSM reconstruction speed by avoiding iterative computation of conventional \textit{improved sparse linear equation and least-squares} (iLSQR) algorithm. To address the same problem, Rasmussen \textit{et al.}~\cite{deepqsm2018} introduced  DeepQSM which is a U-Net based architecture for fast and artifact-free QSM reconstruction without requiring explicit regularization parameters and manual parameter tweaking.

The $\QN$ model is trained based on a unique dataset consisting of $99$ subjects, $52$ HHs and $47$ HCs. When it comes to training ML/AI models, this is a small dataset and as such a challenging task. To address this challenge, different methodologies are used, i.e., transfer learning based on pre-trained weights; data augmentations; the $2$-stage architecture design, and; use of image embedding.  Given the small dataset, the proposed $\QN$ architecture, which is designed based on the integration of the above-mentioned methodologies, achieved relatively high accuracy. As the model looks at the whole image or the whole ganglia area and not a very small manually extracted part, the results are very promising for the classification problem at hand.

Fig.~\ref{fig:CAM_full_size} and Fig.~\ref{fig:CAM_crop} show the heat maps obtained through the Class Activation Map (CAM)~\cite{cam2016} method; red represents the more interesting regions for the $\QN$ to pay attention during the classification task, while darker colors represent less informative regions. Heat maps are meant to provide some form of explanation to the AI model's decisions. More specifically, the heat-maps represent the region of interest for the model to extract differentiating information from the images. the red color represents the more interesting regions for the $\QN$, while darker colors represent less informative regions. Intuitively speaking, these show the areas of the input where the model put more attention and based on which the model mainly made its decision.
In Fig.~\ref{fig:CAM_full_size}, heat maps show the basal ganglia as playing a more important role in the network findings. As shown in Fig.~\ref{fig:CAM_full_size}, by visualizing the heat maps resulted from models trained on full-size images, we realized that the main focus of the network is on the basal ganglia region of the brain. This is technically more interesting because nuclei of the basal ganglia are the main regions of iron accumulation in the brain, and the model is trained on its own to pay more attention to this region. This illustration opens a new horizon in our research path, i.e., taring models by using roughly cropped regions of basal ganglia images as input to the models. As shown in Tables~\ref{table1} and~\ref{table2}, cropping the input images, improves the performance for all models. Heat maps resulted from CAM methods are for cropped basal ganglia regions are illustrated in Fig.~\ref{fig:CAM_crop}. We would like to clarify that our deep learning-based solution does not necessarily require information about the exact nuclei of the basal ganglia boundaries and location. In other words, a rough estimation of the location of the basal ganglia is enough to determine the bounding boxes and accordingly crop the image.

Generally speaking, the $\QN$ model can achieve better results with more variations of the training data. We used data augmentation to assist in improving the generalizability and robustness of the $\QN$ model. Data augmentation is an effective approach to increase the size of the data in an inexpensive fashion. Diversity of MRI machines and their configurations or even the utilized method to reconstruct QSM, R2*, and T1-Weighted images from raw MR data can affect the quality of the reconstructed images, e.g., different contrast and brightness of images. One of the positive aspects of STAGE is that if you have multiple systems and field strengths, then it can level the playing field because it is vendor agnostic. Therefore, we can acquire consistent data from multiple systems. One direction for future research is to use more of the maps from STAGE, and integrating an automatic segmentation approach within the proposed $\QN$ framework.

The main objective of supervised-ML algorithms is building a system/model that can identify what is the correct answer by training on a large number of sample pairs (i.e., input-output pairs) to improve the decision-making process. The main strength of ML approaches, in particular, DNN techniques is that they can learn ``on their own''. Deep learning, as an advanced ML technique, allows researchers to model non-linear decision surfaces through investigating very high dimensional spaces such as MR images. Traditionally, for diagnoses, human experts use cut-off values as dividing points of the continuous  measurements to indicate that someone has the condition of interest or not, while this is not always applicable. Therefore, using DNN models as an expert adviser or even a decision-maker system about the patient (as we have done in this study) offers promising new ways to improve healthcare services. This is mainly due to the fact that DNN models encompass a higher number of factors/parameters than humans and do not depend on a set of explicitly described rules.
In this study, we explored the potential of DNN methods to analyze high dimensional MRI dataset to categorize individuals in two classes of HC and HH. It is worth mentioning that such differential diagnosis task through visually evaluating QSM and R2* images is very challenging and almost impossible, while DNN achieves this importance through processing data and spotting the hidden patterns in the dataset. By developing more reliable prediction models, the goal is that earlier intervention will help to halt the progression of chronic and costly diseases.

Our future plan is to study the contribution of each scan, i.e., QSM, R2*, and T1W to the final prediction. Furthermore, investigating different DNN architectures like 3D convolutions or Vision Transformers could be another direction for future research. We would like to point out that although using 3D convolutions might improve the overall performance, it comes with extra parameters resulting in more computational and processing costs. Moreover, to train a model composed of 3D convolutions or Transformers, a very larger dataset is required. Albeit, in general, increasing the number of patients leads to a better prediction performance in the DNN-based models, which we leave to investigate in future studies.

\bibliography{sample}

\section*{Acknowledgment}
This work was partially supported by the Natural Sciences and Engineering Research Council (NSERC) of Canada through the NSERC Discovery Grant RGPIN-2016-04988.

\section*{Data Availability}	
The datasets generated and/or analyzed during the current study are not publicly available due to the confidentiality restrictions imposed by the approved ethics of the study.

\vspace{-0.1in}
\section*{Author Contributions Statement}
S.Z. and E.R. implemented the deep learning models and performed the evaluations, S.Z. E.R. drafted the manuscript jointly with A.M.; S.S. and M.S.J. supervised the clinical study, data collection, and labeling process and edited the manuscript; S.K.S., S.G., and E.M.H. contributed to sequence generations and reconstructions; S.S., M.S.J., E.M.H., A.A. and A.M. directed and supervised the study. All authors reviewed the manuscript.

\vspace{-0.1in}
\section*{Additional Information}
\textbf{Competing Interests}: Authors declare no competing interests.

\end{document}